\pdfoutput=1
\documentclass[11pt]{article}

\usepackage{acl}

\usepackage{times}
\usepackage{latexsym}

\usepackage[T1]{fontenc}
\usepackage[utf8]{inputenc}

\usepackage{microtype}

\usepackage{graphicx}
\usepackage{caption}
\usepackage{subfigure}
\usepackage{tabularx,booktabs}
\usepackage{threeparttable}
\newcolumntype{P}[1]{>{\centering\arraybackslash}p{#1}}
\newcolumntype{M}[1]{>{\centering\arraybackslash}m{#1}}

\usepackage{makecell}

\usepackage{algorithm}
\usepackage{algpseudocode}
\usepackage{amssymb}
\usepackage{amsmath}
\DeclareMathOperator*{\argmax}{argmax} % thin space, limits underneath in displays
\usepackage{hyperref}

\title{Learning a Better Initialization for Soft Prompts via Meta-Learning}

\author{Yukun Huang$^*$ \\
  Columbia University\\
  \texttt{yh3310@columbia.edu} \\\And
  Kun Qian\thanks{\ \ Equal Contribution} \\Columbia University\\
  \texttt{kq2157@columbia.edu} \\\And
  Zhou Yu\\Columbia University\\
  \texttt{zy2461@columbia.edu}}

\begin{document}
\maketitle
% \qk{1."-->``}
% \qk{white space before parenthesis}
% \qk{~\citet{}}
% \qk{\textit for labels}
% \qk{fonts in figures}
% \qk{a table of examples of all seven datasets in appendix}
% \qk{normalize pre-training and pre-trained}
% \qk{table --> Table}
% \qk{1000-> 1,000}
% \qk{continuous --> soft}
% \qk{advantage}
% \qk{however, we found in few shot setting, it does not perform stably and well}
% \qk{"-->''}
% \qk{firstly-->first}
% \qk{unnecessary words: can, need, in terms of ...}

\begin{abstract}
Prompt tuning (PT) is an effective approach to adapting pre-trained language models to downstream tasks.
Without a good initialization, prompt tuning doesn't perform well under few-shot settings. So pre-trained prompt tuning (PPT)~\citep{Gu2022PPTPP} is proposed to 
initialize prompts by leveraging pre-training data.
We propose \textbf{MetaPT} (\textbf{Meta}-learned \textbf{P}rompt \textbf{T}uning) to further improve PPT's initialization by considering latent structure within the pre-training data.
Specifically, we introduce the structure by first clustering pre-training data into different auxiliary tasks with unsupervised methods. Then we use these tasks to pre-train prompts with a meta-learning algorithm. Such a process can make prompts learn a better initialization by discovering commonalities among these auxiliary tasks. 
We evaluate our method on seven downstream tasks. Our MetaPT achieves better and more stable performance than the state-of-the-art method.
%Our results demonstrate that our initialization requires only 10\% of training data to achieve comparable performance with xxx\qk{SOTA method}.
%
%and outperforms baselines by 2\% with same amount of training data.

% As a result, we train the soft-prompt to be easy to fine-tune. We demonstrate that our meta-learned prompt tuning could outperform pre-trained prompt tuning and full-model tuning.
\end{abstract}

\section{Introduction}

\textbf{Pre-trained language models} (PLMs) (e.g. BERT~\citealp{devlin-etal-2019-bert}; T5~\citealp{Raffel2020ExploringTL}; GPT3~\citealp{Brown2020LanguageMA}) have demonstrated outstanding performances in various downstream NLP tasks.
\textbf{Full-model tuning} (FT) and \textbf{prompting} are two methods that leverage PLMs for downstream tasks. 
% [HOW FT uses PLM], [How promoting use PLM]. 
FT adapts PLMs to downstream tasks by introducing task-specific training objects and fine tuning all parameters of PLMs. Instead of tuning the entire PLMs, prompting methods probe the knowledge in PLMs by templates (i.e. \textbf{prompts}) to solve downstream tasks.
% [FT vs. Prompting]. 
FT shows state-of-art performance in most scenarios. But as size of pre-trained language model increases, fine tuning and then storing the whole model parameters would be quite expensive. In contrast, prompting doesn't need to tune the entire PLMs, which makes it parameter-efficient. Therefore, prompting gradually becomes an alternative solution for FT to utilize large-scale PLMs. 

% [Prompting has two types...] 
There are two types of Prompts: \textbf{hard prompts} and \textbf{soft prompts}. Hard prompts are human-designed discrete tokens while soft prompts are continuous embeddings of language models. \citet{brown2020language} first introduce the concept of hard prompts. They find that behaviors of GPT3 could be modulated by text prompts. For example, we want to know the sentiment of the sentence ``I love this movie''.  We could add a template  ``Overall, it was a <mask> movie'' at the end of the sentence. So the input sentence would be ``I love this movie. Overall, it was a <mask> movie''. The hard prompt forces the PLMs to predict the mask token as well as the sentiment of the sentence. 
Soft prompts are templates with their own tunable parameters and perform prompting directly into the continuous embedding space of the model. Soft prompts could be directly trained by data from downstream tasks, while hard-prompts have to rely on costly human specialists to design templates for each task. Moreover, hard prompts have to be human-interpretable natural language, whereas soft prompts can be the continuous embeddings of the model with more representation ability~\citep{Liu2021PretrainPA}. 
% Furthermore, soft prompts break the limitation of prompts being human-interpretable natural language and provide better representation ability.

% [prompt tuning]
\textbf{Prompt tuning} (PT)~\citep{lester-etal-2021-power} is a simple but effective prompting method that combines hard prompts and soft prompts together. PT adds a series of tunable tokens at the beginning of the sequence as soft prompts and also adds human designed template to the sequence as hard prompts. When adapting PLMs to downstream tasks, PT freezes all parameters of PLMs and only trains the soft prompts. In that way, features of downstream tasks will be learned without breaking the inner structure of PLMs.
% [initialization is important]
PT achieves comparable performance to full-model tuning with sufficient data. But it performs poorly under few-shot settings due to its sensitivity to the initialization of soft prompts. This disadvantage significantly affects the practical application of PT.

% [ppt proposed to solve that but not good enough]
\textbf{Pre-trained prompt tuning} (PPT)~\citep{gu-etal-2022-ppt} is proposed to adapt prompt tuning to few-shot settings. PPT pre-trains soft prompts with self-supervised pre-training tasks and then apply pre-trained prompts to few-shot downstream tasks. PPT generally groups all text classification tasks into different formats and designs a self-supervised pre-training task for each format to pre-train prompts. PPT demonstrate its effectiveness when using large-scale PLMs.

% [PPT limitation and why]
However, PPT still has limitations, as it mixes all pre-training data together and treats each data point equally. %
Since PPT updates prompt parameters at every data point, it learns more about the specific feature of each data point rather than general features of the entire task.
% , as it would over-fit on the pre-training tasks. 
% Since pre-training task and downstream task are different, not all knowledge learned from pre-training data is applicable to downstream tasks. % because these general features are shared by both pre-training task and downstream task.
%
% In fact, some general features are more important than other features in pre-training data because these general features are shared by both pre-training task and downstream task.
%
% These general features would contribute significantly to adapting the model to downstream tasks. 
%
However, knowledge learned from pre-training data is not necessarily applicable to downstream tasks because pre-training task and downstream task are different.
As a result, PPT retains too much redundant information only relevant to the pre-training task in the initialization of soft prompts, which consequently impedes model performance on downstream tasks. 
% Therefore, general features shared by the whole dataset are more important.
% However, PPT doesn't fully explore the inner structure of pre-training data and treats each data sample equally. 
%
% As a result, PPT loses many important general features and retain too much redundant information only relevant to the pre-training task in the initialization of soft prompts, which consequently impedes model performance on downstream tasks. 
% The bigger in difference between the pre-training task and the downstream task, the worse the performance of PPT.
% [we propose ....]
% [use cluster to learn inner structure and metalearning to utilize the inner structure to learn the commonality between tasks]

To obtain a better initialization for soft prompts, we incorporate meta-learning into prompt tuning. We propose an innovative unsupervised method to create meta-learning tasks for prompts and then introduce a model-agnostic meta-learning method to pre-train prompts. By our unsupervised clustering method, latent structure of pre-training data is represented by the distribution of meta tasks. Through meta-learning, general features are incorporated to the initialization of the soft prompts. Our meta-learned prompts achieve faster and more stable adaptation to downstream tasks. We named our method \textbf{Meta}-learned \textbf{P}rompt \textbf{T}uning ``\textbf{MetaPT}''. Our experiments show that MetaPT outperforms full-model tuning and pre-trained prompt tuning on the base-size model. 

\section{Related work}
\textbf{Hard Prompts} The basic idea of hard prompts is to use a discrete human-designed natural language prompt to query a language model. Hard Prompt first show its effectiveness in \citet{Brown2020LanguageMA}. They discover that a frozen GPT-3 model's behavior could be modulated by text prompts. After that, much effort is made to automatically search for prompt templates in a discrete space. \citet{jiang-etal-2021-know} apply a mining-based method to automatically search for suitable templates in a given task. \citet{Yuan2021BARTScoreEG} and \citet{haviv-etal-2021-bertese} perform paraphrasing to prompts with different methods. Basically, they both paraphrase an already constructed prompt as a seed prompt and then paraphrase it into a set of candidate prompts and select the prompt with the best performance from this set. \citet{shin-etal-2020-autoprompt} implement gradient-based search over actual tokens to find templates with downstream application training samples. \citet{gao-etal-2021-making} and \citet{BenDavid2022PADAEP} turn prompt templates searching into task-generation task and use text-generation model T5 to generate prompts.\\ \\
\textbf{Soft Prompts} There are also methods exploring soft prompts which incorporate prompting directly into the continuous embedding space of the model. Prefix tuning \citep{li-liang-2021-prefix} is a method that freezes the model parameters and tunes the prefix activations prepended to each layer in the encoder stack.  \citet{lester-etal-2021-power} propose a further simplified approach called prompt tuning, which only tunes the additional tunable tokens prepended to the input text. P-tuning v2 \citep{Liu2021PTuningVP} adapted the idea of prompt tuning by adding prompts in different layers as pre-fix tokens rather than only the input embedding. Its performance can be comparable to full-model tuning across both scales and tasks. 
Though the above soft prompt related methods perform well with sufficient training data, they all become much worse under few-shot learning settings.  Pre-trained prompt tuning \citep{Gu2022PPTPP} is introduced to enhance the performance of prompt tuning when training data is limited. They group typical classification tasks into three formats and create one self-supervised pre-training task for each format. Pre-trained by these self-supervised tasks, prompt tuning could reach or even outperform full-model tuning methods under few-shot settings.\\ \\
\textbf{Meta Learning}
Meta-Learning, or learning about learning, aims to improve the learning algorithm itself. One popular meta-learning framework would be Model-Agnostic Meta-Learning(MAML) proposed by \citet{Finn2017ModelAgnosticMF}. MAML could be directly applied to any learning problem and leads to great performance with a small amount of training data. MAML-related approaches are adopted in various NLP tasks. \citet{dou-etal-2019-investigating} treat some high resources datasets in GLUE \citep{wang-etal-2018-glue} as auxiliary tasks and use the rest of them as target tasks to test the performance. \citet{qian-yu-2019-domain} applies MAML to an end-to-end dialog system for domain adaption and then \citet{Qian2021ASA} incorporates a meta-teacher model to optimize the domain adaption process.

\section{Background}
% \qk{in this section we gonna talk about first..... then .....}
In this section, we introduce basic knowledge about prompt tuning. Following \citet{lester-etal-2021-power}, we turn all the text classification tasks into text generation tasks. We first pre-process input samples to adjust to prompt tuning setting and then formulate the idea of prompt tuning.
\subsection{Pre-processing}
 Pre-processing is necessary to adjust the input sample to text-to-text form. Specifically, there are two steps to pre-process the input samples.\\ %\qk{not intuitive, give example} \\ 
\textbf{Label Mapping} We first map labels to our pre-defined concrete tokens.
%For label mapping, the method to map labels of specific tasks to tokens is called a verbalizer. 
%Different verbalizers significantly influence the performance of the pretrain language model. 
Taking a 5-class sentiment classification task as an example, the original labels are numbers from 0-4 which denotes the sentiment intensity. We map $0 \to ``terrible"$, $1 \to ``bad"$, $2 \to ``maybe"$, $3 \to ``good"$, $4 \to ``great"$ and then directly use these mapping results as label tokens. Since pre-trained language model is trained on natural languages, mapping labels to real words usually leads to decent performance \cite{gu-etal-2022-ppt}. \\
\textbf{Prompt Adding} In the second step, we add prompts to the input sentence. we implement a hybrid prompt \citep{Gu2022PPTPP} which combines soft prompts and hard prompts together to achieve the best performance.
The prompt we applied is designed with the following template: a set of soft prompts embedding is prepended to the input sequence features, and then the manually desigend hard prompt ``It was $\langle X \rangle$ . '' would be added at the end of the input sentence, where $\langle X \rangle$ denotes the mask token in the pre-trained language model. \\
\subsection{Formulation}
Different from full-model tuning which tunes parameters of the entire pre-trained language model, prompt tuning only modifies the parameters of prompts prepended to the input sentence.
Fine-tuning process of adapting pre-trained language model to a downstream classification task $\mathcal{T}^{down}$ could be represented by optimizing the following log-likelihood objective:
\begin{align*}
    \argmax_{\theta}\sum \log p(y|x,\theta)
\end{align*}
where  $(x, y)$ is an input sample in $\mathcal{T}^{down}$, $x$ is an input sentence and $y$ is its label. \\
Prompt tuning freezes parameters of language model and only tunes soft prompts $\boldsymbol{P}$. Mathematically, hybrid prompt tuning could be represented by optimizing the following log-likelihood objective: 
\begin{align*}
    \argmax_{\boldsymbol{P}}\sum \log p(\langle X \rangle = z|[\boldsymbol{P};H(x)];\boldsymbol{P})
\end{align*}
where $z$ is the concrete token mapped from label y, $H(x)$ denotes fitting input sentence $x$ to a hard prompt template and $[\boldsymbol{P};H(x)]$ denotes prepending soft prompts $\boldsymbol{P}$ to the beginning of input sequence.

Pre-trained prompt tuning~\cite{Gu2022PPTPP} proposes to pre-train soft prompts $\boldsymbol{P}$ on rich-resource pretraining dataset $\mathcal{D}^{pre}$, in order to improve the performance on downstream tasks $\mathcal{T}^{down}$  under few-shot settings. 
Instead of directly training prompts on pre-training data $\mathcal{D}^{pre}$, we propose meta-learned prompt tuning to further enhance the performance of prompt tuning under few-shot settings. 
Pre-training data $\mathcal{D}^{pre}$ is first divided into different auxiliary meta tasks $\mathcal{T}^{meta}$  by unsupervised methods and then prompts are trained during and a model-agnostic meta-learning phase on these meta tasks.

\section{Meta-learned Prompt Tuning}
\label{sec:Meta-learned Prompt Tuning}
% \qk{this is a pipeline in terms of training our model, in sec4.1 we gonna ..  in sec4.2 we gonna ....}
% We aim to train prompts that can achieve rapid adaptation to downstream tasks, which would be a great fit to solve few-shot learning problems. To accomplish this, the prompts are trained with model-agnostic meta-learning on a set of tasks. 
%
In this section, we describe our model training pipeline.
% As shown in figure \ref{fig:flowchart}, there are mainly four steps in our approach. \\
%
% In \hyperref[sec:Constructing Pre-training Data]{section 4.1}, 
We first describe the process of gathering pre-training data for meta-learning. 
% We are gonna talk about our two methods to construct pre-training data in \hyperref[sec:Constructing Pre-training Data]{section 4.1}  \qk{connection}
% It could be either a high resource dataset similar to our downstream datasets or a large open-domain corpus with pseudo labels created unsupervisedly.
% In \hyperref[sec:Designing Meta-Learning Tasks]{section 4.2}, 
Then, we introduce different unsupervised methods to cluster pre-training data into different groups as auxiliary tasks for meta-learning.
% We consider each class as an auxiliary task for meta-learning. We introduce two different unsupervised clustering  methods 
% In \hyperref[sec:Prompt-MAML Algorithm]{section 4.3}, 
Finally, we describe our prompt-MAML algorithm to train and find a good initialization for soft prompts.

We aim to encode more general features shared by pre-training data and downstream tasks to the initialization of prompts so that the model could adapt faster to downstream tasks. 
% With general features shared by meta tasks and down-stream tasks, prompts could achieve faster adaption to downstream tasks.

\begin{figure*}[htbp]
\centering
\includegraphics[width=1\textwidth]{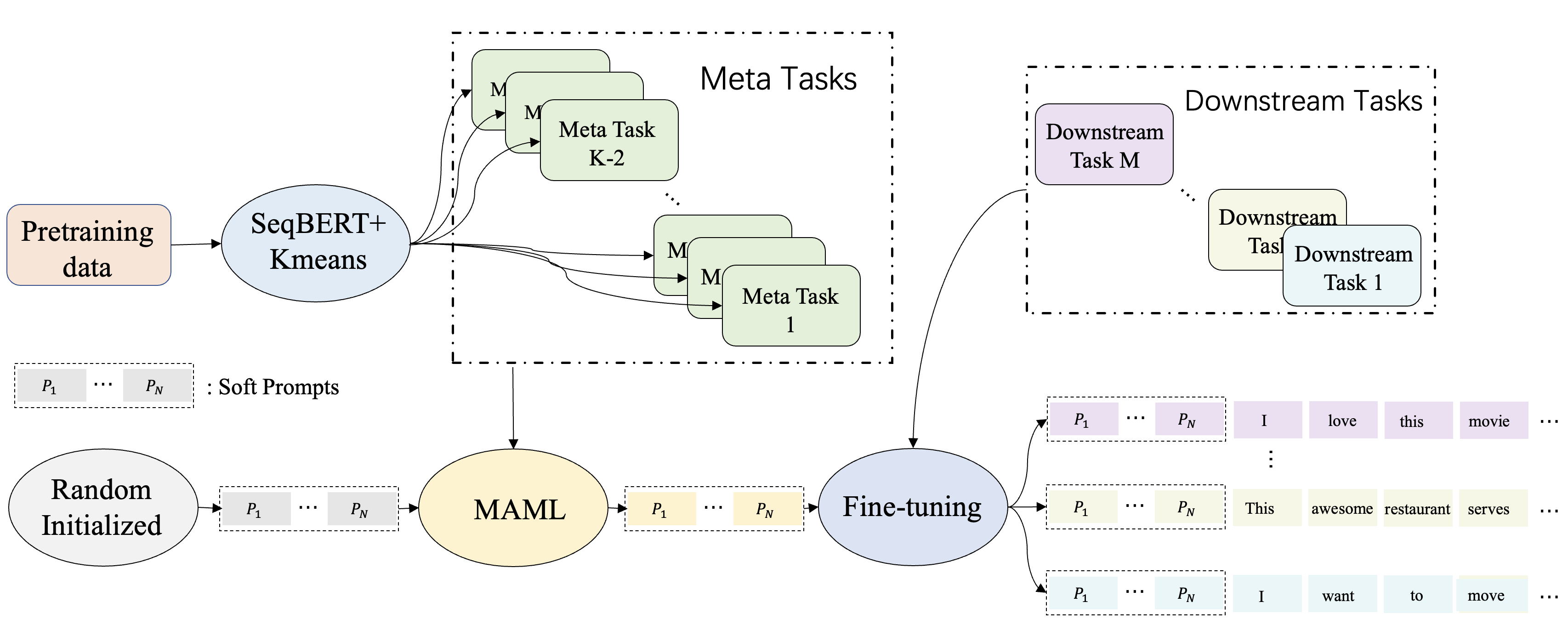}
\caption{Pipelines of meta-learned prompt tuning. First, we prepare pre-training data used for meta-learning. Second, we cluster pre-training data into different groups as auxiliary tasks for meta-learning. Then, we train prompts with model-agnostic meta-learning algorithm. Finally, we evaluate meta-learned prompts on downstream tasks. }
\label{fig:flowchart}
\end{figure*}

\subsection{Constructing Pre-training Data}
\label{sec:Constructing Pre-training Data}
% \qk{where to get the pre-training data}
% \qk{do not put purpose at the beginning} \qk{there are two ways to construct our data, first, .. second ..., the advantage of the first one is .. the advantage of second.... if .. use the first,.in our experiemnt setting, we .....}
There are mainly two methods to construct pre-training data for our algorithm. The first one is to directly choose a high resource dataset similar to the target task as pre-training data.  The second one is to create pre-training data unsupervisedly. In our experiment setting, we focus on sentiment classification tasks.  It's worth mentioning that for sentiment classification task, we still couldn't avoid relying on an existing dataset to create pseudo labels in our second method. The main difference between the first method and the second method is that the pre-training data in the method only covers one domain while the pre-training data created by the second method is an open domain. Besides, from the second method we could create unlimited data as long as we are able to collect enough large corpus from open web. 

For the first method, we directly treat a 5-class sentiment classification large-scale dataset Yelp5 \citep{Zhang2015CharacterlevelCN} as pre-training data. For the second method, we adopt the method proposed by \citet{Gu2022PPTPP} to create pseudo labels for sentences from a large open-domain corpus. We train another model to annotate pseudo labels for the sentences in a large corpus. 

\subsection{Designing Meta-learning Tasks}
\label{sec:Designing Meta-Learning Tasks}
% \qk{why to do clustering}
% [motivation, to cluster for exploring inner structure]
% The question here is how could we find so many tasks for meta-learning.
% The most straightforward idea is to find lots of other similar downstream tasks as meta-learning tasks.
% Though the data would be of high quality, there are mainly two drawbacks of this approach. Firstly, given that the size of different datasets naturally varies across different tasks, we have to throw samples in the high-resource dataset to balance the size of different datasets. Secondly, similarities between our target downstream tasks vary and there could be a task which is much more similar to our target downstream task than others. During the meta-learning, the task which is most similar to the main task will become dominant. As a result, other tasks less similar will have a negative impact on initialization, and meta-learning will fail.
After constructing pre-training data in \hyperref[sec:Constructing Pre-training Data]{section 4.1}, we group the data into different clusters as auxiliary meta tasks used for meta-learning. We propose two unsupervised methods to separate the pre-training data into several meta tasks. By these unsupervised clustering methods, latent structure within pre-training data is revealed. Based on that structure, prompts could learn to incorporate some common internal features to the initialization through meta-learning.  With such general information encoded in the initialization of prompts, the model can achieve great performance with limited training data from downstream tasks. 
\subsubsection{Kmeans Tasks}
In this method, we group pre-training data into different classes by K-means clustering. Assume pre-training data we get from \hyperref[sec:Constructing Pre-training Data]{section 4.1} is 
\begin{align}
    \mathcal{D}^{pre} = \{ \boldsymbol{x,y} \} 
\end{align}
We first implement sentence-BERT(\citealp{reimers-gurevych-2019-sentence}) to derive semantically meaningful sentence embeddings from pre-train data samples. 
\begin{align}
    s = SeqBERT(x)
\end{align}
Then we apply unsupervised K-means to cluster pre-training data into different classes according to their sentence embeddings. We first set K to the number of clusters, then we generate K different meta tasks from pre-training data:
\begin{align}
    \{\mathcal{T}_i, \ \  i = 1...K\}  &= Kmeans(\{ \boldsymbol{x,y} \},\boldsymbol{s}, K) \\
    \mathcal{T}_i &= \{\boldsymbol{x}^{(i)},\boldsymbol{y}^{(i)},i\} 
\end{align}
We want to avoid prompts from learning too much redundant semantic information that only relevant to pre-training data, but learning how to extract information from language models for sentiment classification.
 % in such types of tasks. 
 Through K-means clustering, samples containing similar sentence embeddings are grouped into the same task. During the meta-learning process, prompts keep the general features of task type (sentiment classification in our case) and throw away some redundant semantic information only relevant to domain of pre-training data. The task type information retained in the initialization is going to play a key role in the subsequent main task few-shot learning. 
%part of semantic information is not learned by prompts, the general features of this type of task are kept in prompts.

\subsubsection{LDA Tasks}
Latent Dirichlet Allocation (LDA) is an alternative way to group pre-training data into different clusters. It is a generative statistical model to automatically group documents into different topics. LDA aims to discover the hidden themes in the collection of data. We apply LDA to group pre-training data into different tasks according to their topics.  Pre-training samples in the same task would have similar themes while samples across different tasks would differ in their themes. By grouping pretraining data into different clusters according to their hidden themes, we hope to eliminate unimportant variations in the topics among pre-training data, while maintaining information related to task type in the prompts.

\subsection{Prompt-MAML Algorithm}
\label{sec:Prompt-MAML Algorithm}
After we get a set of meta tasks $\mathcal{T}$ obtained via an unsupervised method, we utilize MAML to learn general features among these meta tasks. We first randomly initialize the parameter of soft prompts $\boldsymbol{P}$. For each meta task $\mathcal{T}_i$, m training samples are sampled from that task. Taking in m samples, the model output $f_{\boldsymbol{P}}$. Then we calculate the average loss $\mathcal{L}_{\mathcal{T}_i}(f_{\boldsymbol{P}})$ of these m samples and temporarily updates soft prompts with gradient descent. 
\begin{align}
    \boldsymbol{{P} _i'} = \boldsymbol{P}-\alpha \nabla_{\boldsymbol{P}} \mathcal{L}_{\mathcal{T}_i}(f_{\boldsymbol{P}})
\end{align}
After optimizing the prompts, we sampled another m samples and calculate the loss with the updated prompts. We add loss for $\mathcal{T}_i$ to total loss and repeat the same process for other meta tasks until we go over all the meta tasks.
Finally, we update the prompts by minimizing the final total loss. 
\begin{align}
\boldsymbol{P} \gets \boldsymbol{P} - \beta \nabla_{\boldsymbol{P}} \sum_{\mathcal{T}_i \sim p(\mathcal{T})}\mathcal{L}_{T_i}(f_{\boldsymbol{P_i'}})
\end{align}
This is a complete process of one-step updates for prompts. We keep optimizing the prompts until the validation accuracy of meta tasks stop growing. The whole algorithm of Meta-learned prompt tuning is shown below.
\begin{algorithm}
\caption{Prompt-MAML}\label{alg:euclid}
\begin{algorithmic}[1]
\State $\mathcal{D}^{pre}$ : Pretrain data
\State $K$:cluster numbers 
\State $\alpha, \beta$:step size hyperparameters
\State $\mathcal{T} =$ UnsupervisedClustering($\mathcal{D}^{pre}, K$)
\State randomly initialize soft prompts $\boldsymbol{P}$ \\
\While{not done}
    \For {$\mathcal{T}_i \in \mathcal{T}$}
        \State Evaluate $\nabla_{\boldsymbol{P}} \mathcal{L}_{\mathcal{T}_i}(f_{\boldsymbol{P}})$ with respect to m samples
        \State compute adapted parameters with gradient descent: ${P} _i' = \boldsymbol{P}-\alpha \nabla_{\boldsymbol{P}} \mathcal{L}_{T_i}(f_P) $
    \EndFor
    \State $P \gets P - \beta \nabla_P \sum_{\mathcal{T}_i \sim p(\mathcal{T})}\mathcal{L}_{T_i}(f_{P_i'})$
\EndWhile
\label{euclidendwhile}
\State \textbf{return} $\theta$\
\end{algorithmic}
\end{algorithm}

\begin{table*}[t]
\centering
% \begin{tabular}{llllllll}
\begin{tabularx}{1.0\textwidth}{XXXXXXXX}
\hline
\textbf{Methods} & \textbf{SST5} & \textbf{SST2 } & \textbf{Amazon5} &\textbf{Amazon2} & \textbf{Sentihood} &\textbf{SemEval$_r$} & \textbf{SemEval$_l$}\\
\hline
FT & $43.57_{\pm2.56}$ & $88.27_{\pm1.03}$ & $48.40_{\pm1.48}$ & $92.35_{\pm0.68}$ & \boldmath$82.11_{\pm1.30}$ &$71.01_{\pm1.16}$ &$62.48_{\pm3.23}$\\ 
PPT & $42.90_{\pm1.08}$ & $87.42_{\pm1.15}$ & $51.15_{\pm1.56}$ & $93.28_{\pm0.21}$ & $80.06_{\pm3.31}$ &$62.04_{\pm3.34}$ &$56.37_{\pm4.11}$\\ 
MetaPT & $45.26_{\pm0.39}$ & \boldmath$89.47_{\pm0.12}$ & $55.47_{\pm0.34}$ & $94.43_{\pm0.08}$ & $80.38_{\pm0.46}$ &$76.93_{\pm1.19}$ &$70.86_{\pm1.95}$\\  
MetaPT$_{(Y)}$ & \boldmath$46.24_{\pm0.42}$ & $87.26_{\pm0.73}$ & \boldmath$58.73_{\pm0.13}$ & \boldmath$95.39_{\pm0.03}$ & $78.27_{\pm1.17}$ &\boldmath$80.72_{\pm0.60}$ &\boldmath$72.32_{\pm0.66}$\\ 
\hline
\end{tabularx}
\caption{\label{main table}
Sentiment classification results on seven datasets. Our methods outperform the two baselines for most of the datasets. The meta-learned prompts trained on pseudo-labeled data (MetaPT) not only achieve higher accuracy than PPT consistently, but also have a more stable performance with lower variance.
}
\end{table*}

\section{Experiments}
\label{sec:experiments}
Our experiments are built on the T5 base model from HuggingFace~\cite{wolf-etal-2020-transformers}.\\
\textbf{Downstream Datasets}  We focus on the sentiment classification tasks. Specifically, the downstream datasets include SST-5~\citep{socher-etal-2013-recursive}, SST-2~\citep{socher-etal-2013-recursive}, Amazon-5~\citep{Zhang2015CharacterlevelCN}, Amazon-2~\citep{Zhang2015CharacterlevelCN}, Sentihood~\citep{saeidi-etal-2016-sentihood}, and SemEval-2016~\citep{pontiki-etal-2016-semeval}. SemEval-2016 has two tasks in different domains: restaurant and laptop. These two tasks are denoted by SemEval$_r$ and SemEval$_l$ respectively. Detailed information of these datasets could be found in Appendix \ref{sec:dataset_examples}. We randomly select 40 samples from original dataset for both few-shot training and validation. \\
\textbf{Pre-training Data} We mainly gathering two different sources of pre-training data here. For the first source, we directly choose Yelp5 as pretraining data. Yelp5 has 650,000 training samples only covering the domain of restaurant. For the second source,  we first train a RoBERTa-base model on Yelp5. Then we randomly sample 10GB of data from OpenWebText \citep{Gokaslan2019OpenWeb} and apply the trained RoBERTa model to annotate labels for the sampled data. We only keep data samples with high confidence and throw away the samples which model is unsure about. After balancing pseudo data, we get 1,000,000 balanced training samples with open domains. These two sources of data correspond to MetaPT$_{(Y)}$ and MetaPT respectively in Table \ref{main table}. For both two sources of data We set cluster number to 10. We implement Sentence-BERT-base model to extract sentence features from pre-training data. Then we apply K-means to group pre-training data into 10 clusters as auxiliary tasks for meta-learning.\\
\textbf{Hyperparameters} Following \citet{lester-etal-2021-power}, we set the soft-prompt as 100 tunable tokens. During the meta-learning phase, we implement Adam with weight decay as the optimizer. We set the learning rate $\alpha$ to 0.08, learning rate $\beta$ to 0.025, the batch size to 4, early stop patience to 6, and the max updating step of MAML to 20000. During the downstream prompt tuning phase, we implement AdamW as the optimizer. We use the linear scheduler with 20 warm up steps to achieve a learning rate of 0.003. We set batch size to 4, the max epoch to 200, and the patience for early stopping to 5. To achieve a reliable result, we run all the experiments for five times with different random seeds and report the mean numbers, along with their deviations.
More experimental arguments of pre-trained prompt tuning and full-model tuning could be found in Appendix \ref{sec:training_settings}.

\section{Results}
\subsection{Main Results}
As shown in the Table~\ref{main table}, we mainly compare the performance of full-model tuning (FT), pre-trained prompt tuning (PPT), meta-learned prompt tuning (MetaPT) on different sentiment classification tasks. We also include results of MetaPT$_{(Y)}$ in the table, which is a variation of MetaPT. Instead of being trained on the pseudo data, MetaPT$_{(Y)}$ is directly trained on Yelp5. According to \citet{Gu2022PPTPP}, the performance of plain prompt tuning lags far behind FT and PPT, we don't include it in our main table. We have three observations from results in Table~\ref{main table}.

% metapt vs. baselines
First,  MetaPT consistently achieves better results than PPT over all seven tasks.
% why
MetaPT outperforms PPT by prevent overfitting on pre-training datasete. Benefitting from MAML algorithm, MetaPT only transfer general features to the initialization of prompts and throw out redudant information.
% metapt is better than ft except % why better
MetaPT also outperforms FT for most of the tasks. Because we utilize the extra pre-training data to train the soft prompt without destroying the inner structure of language model.
% except and why except
However, full-model tuning achieves the best score on the Sentihood dataset because the examples in this dataset are simple and easy to learn for full-model tuning. From the example shown in Appendix \ref{sec:dataset_examples}, the input of Sentihood is shorter than other datasets, and the words appeared in this sentihood is more commonly used in daily life. Also, its labels only has 2 classes \textit{good} and \textit{bad}. Due to above three reasons, we believe that training samples of Sentihood are close to what language model learned through pre-training and  would cause less destruction to inner structure of language model through full-model tuning. Therefore, full-model tuning can quickly learn the Sentihood task with a few data points. 

% metapt y vs metapt p
Second, MetaPT$_{(Y)}$ also outperforms PPT and FT on most tasks. And MetaPT$_{(Y)}$ achieves even better results than MetaPT on some tasks.
We look into the results on SST-5 and we find that the recall score of MetaPT is much lower than MetaPT$_{y}$ on the label \textit{terrible} (0.22 vs. 0.61). We believe this is caused by some pseudo data samples with poor quality. When creating pseudo labels, the classifier trained by Yelp5 tends to annotate text with the label \textit{terrible} with high probability, even though the text is irrelevant to \textit{terrible} from the human perspective. For example, ``Both parties agreed to have 1,410 troops located in juba, lueth said, and there are already 1,370 opposition soldiers in the capital.'' is considered as \textit{terrible} with high confidence by classifier while we don't discover any sentiment here. This tendency results in the low quality of pseudo data with the label \textit{terrible} which confuses the MetaPT during MAML. 
% metapt general
% according example 
% metapty general
Even though Yelp5 covers only restaurant domain, MetaPT$_{(Y)}$ still shows great performance when adapting to sentiment tasks in other domains. This suggests that our MetaPT is able to learn the general features, which enable the prompt be easily generalized to other domains. 

% stability
Finally, MetaPT is much more stable than both PPT and FT. Most methods would suffer from high variance under few-shot setting due to their sensitivity to different training samples. The performance of FT varies tremendously when we select different few-shot samples for FT. However, MetaPT shows remarkable stability facing different training samples. The standard deviation of MetaPT is much lower than both FT and PPT across all downstream tasks. This phenomenon indicates that MetaPT is more robust under few shot settings.
\begin{figure}[h]
\begin{center}
  \includegraphics[width=0.45\textwidth]{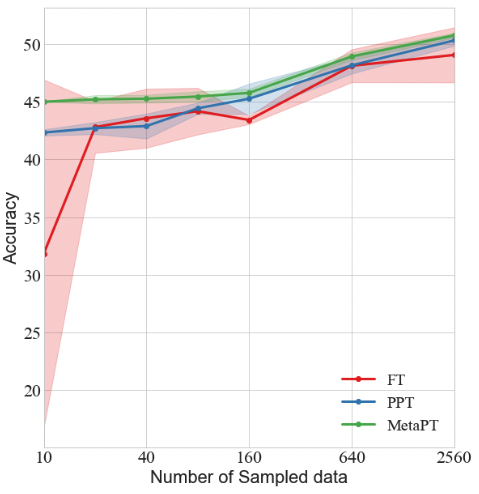}
  \caption{Performance comparison among FT, PPT, MPT on SST-5 as the number of training samples increase from 10-2,560}
  \label{fig:exp1}
\end{center}
\end{figure}

We show the tendency of how performance of FT, PPT, MetaPT varies when the number of training samples increases on SST-5. As shown in Figure \ref{fig:exp1}, when the number of training samples grows from 10 to 2,560, MetaPT is consistently better than PPT and FT, while PPT also has a small advantage over FT. It should be noted that, the full-model tuning method will eventually catch up with other two prompt tuning methods as the number of training samples keep increasing. All three methods will converge to similar performance when training data is sufficient.

\begin{figure}[htbp]
    \centering
    \subfigure[pre-training data size]{ \label{fig: data size}
        \includegraphics[width=1.4in]{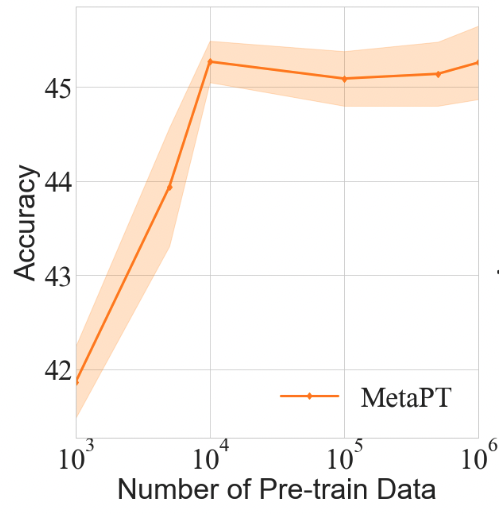}
    }
    \subfigure[number of clusters]{ \label{fig: cluster number}
	\includegraphics[width=1.4in]{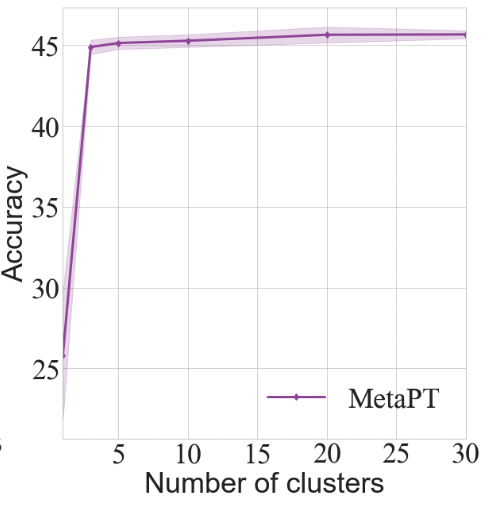}
    }
    
    \caption{Analysis about MetaPT. (a)  The performance of MetaPT varies when the number of pretraining samples change from 1,000 to 3,000,000 (b) The performance of MetaPT varies when the number of  Meta Tasks varies from 3 to 30 }
\end{figure}

\begin{figure*}[htbp]
\centering
\includegraphics[width=1\textwidth]{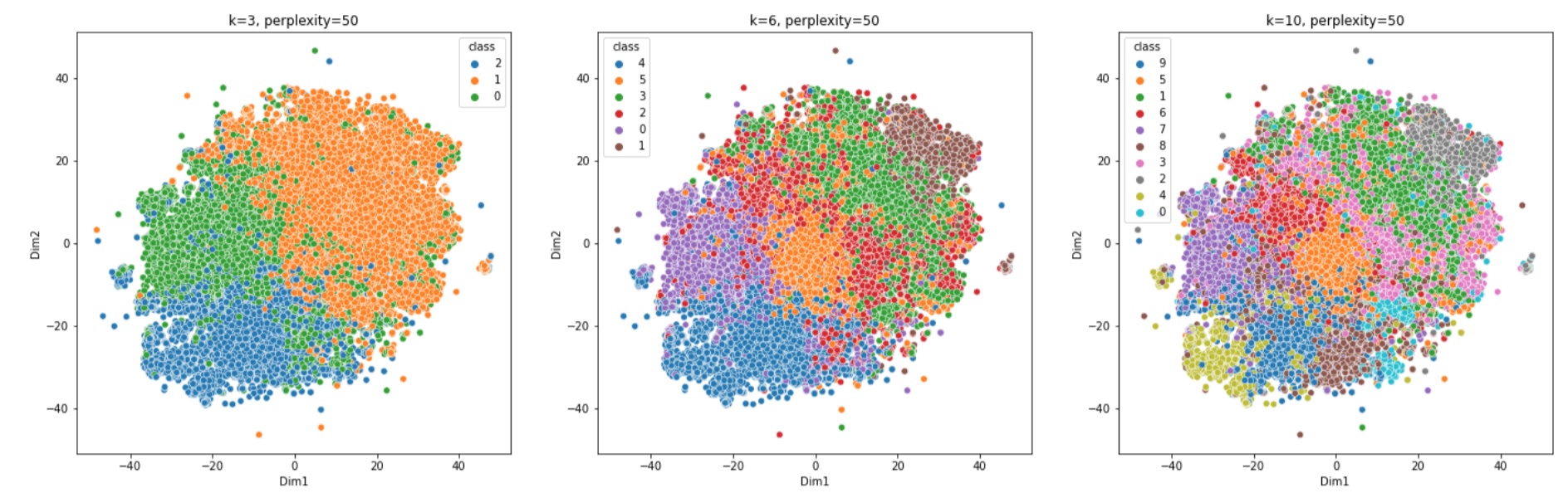}
\caption{tsne of kmeans meta tasks clustering results. Cluster number K equals to 3, 6, 10 respectively }
\label{fig:tsne}

\end{figure*}

\subsection{Ablation Study}
In this section, we discuss how much pre-training data we need to obtain a good result, how the method of clustering affects the performance and how the number of clusters influences results. All the experiments are evaluated on SST-5.\\
\textbf{Scale of pre-training data} 
% Given that the pre-train data size we could collect naturally varies in different types of tasks, it's possible that we couldn't find such abundant pre-training data in some other tasks. When the data size is limited, the data distribution may not well represent general features, which will result in the poor results of our clustering methods. Thus, it's necessary to explore how much pre-training data is needed in our meta-learning algorithm. We want to minimize the pre-training data we used without undermining the performance of our model. 
We want to minimize the pre-training data we used without undermining the performance of our model. We set the number of pre-train samples from 1,000 to 1,000,000. And then we implement Kmeans to cluster them into 10 classes for meta-learning. As the Figure \ref{fig: data size} shows, the accuracy grows rapidly when the number pre-training data increases from 1,000 to 10,000. After 10,000 training samples, the performance does not change much as the number of training samples increases. This result suggests that more pre-training data samples are not necessarily along with better performance in our method. When the size of pre-train data reaches the level of 10,000, it is enough for our model to get acceptable performance.\\
% To explore how the unsupervised clustering would influence the performance of MetaPT, we investigate it from two perspectives: one is how the method of clustering affects the performance and another is how the number of clusters influences results. Both two experiments are evaluated on SST5. \\
\textbf{Methods of clustering} We design four different methods of clustering to get meta tasks. They are K-means clustering, LDA clustering, random clustering, and label clustering. For K-means clustering, we apply a Sentence-BERT-base model to extract sentence features from pre-training samples and then implement K-means to cluster these samples into different groups according to their sentence features. For LDA clustering, we group training samples according to the hidden themes extracted by the LDA topic model. For the random method, we randomly split the total pre-train dataset into different groups. For the label method, we cluster the data samples with the same label into the same group. When the clustering number is large than the number of labels, we just randomly split samples with the same label into more groups.We fixed the cluster number as 10 in this experiment. 
From the result shown in Table \ref{clustering methods} we notice that K-means clustering is most effective and LDA is next to it. When we cluster randomly or according to their labels, the performance of MetaPT degrades to the same level as PPT.\\
\begin{table}
\centering
\begin{tabular}{ll}
\hline
\textbf{Methods} & \textbf{Accuracy} \\
\hline
K-means & 46.24$\pm$0.42 \\ 
LDA & 44.10$\pm$0.83 \\
random & 42.95$\pm$1.05 \\
label &  42.84$\pm$0.76\\ 
PPT &  42.89$\pm$1.08\\
\hline
\end{tabular}
\caption{\label{clustering methods}
Performance of different clustering methods on SST5. ``Kmeans'' and ``LDA'' denote methods we mentioned in \hyperref[sec:Designing Meta-Learning Tasks]{section 4.2}. ``Random'' denotes the method which randomly splits the total pre-training data into different groups. ``Label'' denotes the method which clusters the data samples with the same label into the same group. Pre-trained Prompt Tuning(PPT) plays the role as a baseline.
}
\end{table}
\textbf{Number of clusters}   We examine the cluster number from 3-30 and compare performance. We fix the clustering method as K-means. As shown in the Figure \ref{fig: cluster number}, the accuracy grows rapidly at first as cluster number increases but later it converges. Considering both effectiveness and efficiency, MetaPT is able to achieve promising results when k=10. We also visualize the result of K-means clustering when cluster number equals to 3, 6, 10 in Figure \ref{fig:tsne}.
From the TSNE of our K-means clusters, we could see that data is well grouped into different clusters according to their sentence embeddings.  After we reduce the sentence features of Different samples to two-dimensionality, different samples in the same clusters are close to each other and  are distinguishable from samples in other clusters, which demonstrates that meta tasks derived from K-means indeed contain useful common latent features.

\section{Conclusions}
In this paper, we present the meta-learned prompt tuning framework. 
Specifically, we propose to cluster pre-training data into different groups to create auxiliary tasks for meta-learning, and then pre-train prompts with the Model-Agnostic Meta-Learning method. 
We explore our method based on the sentiment classification task and evaluate our meta-learned prompts on seven downstream datasets under a few-shot setting. 
The results demonstrate that meta-learned prompt tuning achieves better performance and stability than the state-of-the-art methods. 
We also conduct ablation study on different sources of pre-training data and different ways to obtain meta tasks. 

In the future, we plan to apply our method to larger pre-trained language model, e.g. T5-xxlarge. 
We also plan to extend our evaluation tasks from sentiment classification to other general natural language processing tasks, e.g. sentence pairing, to explore the generalizability of our method.
We hope that our work stimulates further research in how to leverage prompts to solve NLP tasks with pre-trained language models.

% \section{Limitations}
% Meta-learned Prompt Tuning (MetaPT) outperforms full-model tuning even on a base size language model only under the premise that the pre-training data is similar to the downstream tasks. In our case, they are both sentiment classification tasks. However, when the type of pretraining dataset is quite different from the downstream dataset, MetaPT is not able to outperform full-model tuning under few-shot settings. 
% A larger language model like T5-xxl is necessary to fully explore the potential of MetaPT under few-shot settings.
% \label{sec:bibtex}

% \section*{Acknowledgements}

% We would like to thank the anonymous reviewers for their suggestions and comments
% Additional elements were taken from the formatting instructions of the \emph{International Joint Conference on Artificial Intelligence} and the \emph{Conference on Computer Vision and Pattern Recognition}.

\bibliography{anthology,custom}
\bibliographystyle{acl_natbib}

\newpage
\appendix
\onecolumn
\section{Training Settings}
\label{sec:training_settings}

We provide detailed training settings used for full-model tuning (FT), pre-trained prompt tuning (PPT). Instead of following \citet{gu-etal-2022-ppt}, we find another set of hyperparameters. Both FT and PPT achieve better performances on T5-base model than results reported in \citet{gu-etal-2022-ppt}. 
\subsection{Full-model Tuning}
We implement AdamW as the optimizer. We apply a linear scheduler with 20 warm up steps and set the learning rate to 0.00003. We set batch size to 4, max epochs to 200. We evaluate results on validation set every epoch and and set the patience for early stopping to 5.
\subsection{Pre-trained Prompt Tuning}
We apply pseudo data created in \hyperref[sec:experiments]{section 5} as pre-training data for PPT. 
During pre-training phase, we implement AdamW as the optimizer. We apply the linear scheduler with 20 warm up steps and set the learning rate to 0.003. We set the batch size to 4 and max epoch to 5 (1,250,000 max steps). We evaluate prompts on validation set every 20,000 steps and set the patience of early stop to 5. 

During downstream prompt tuning phase, we adopt the same training setting as downstream prompt tuning in meta-learned prompt tuning. We implement AdamW as the optimizer. We use the linear scheduler with 20 warm up steps and set the learning rate to 0.003. We set batch size to 4, the max epoch to 200, and the patience for early stopping to 5.

\section{Dataset Examples}
\label{sec:dataset_examples}

Here we provide detailed information and examples for all the datasets we used. Pre-training dataset includes Yelp5~\citep{Zhang2015CharacterlevelCN}. The downstream datasets include SST-5~\citep{socher-etal-2013-recursive}, SST-2~\citep{socher-etal-2013-recursive}, Amazon-5~\citep{Zhang2015CharacterlevelCN}, Amazon-2~\citep{Zhang2015CharacterlevelCN}, Sentihood~\citep{saeidi-etal-2016-sentihood}, and SemEval-2016~\citep{pontiki-etal-2016-semeval}. SemEval-2016 has two tasks in different domains: restaurant and laptop. These two tasks are denoted by SemEval$_r$ and SemEval$_l$ respectively. Domains, number of classes and examples of all datasets are shown in Table~\ref{table:dataset_examples}.

\begin{table*}[htbp]\small
    \centering
    % \resizebox{1\linewidth}{!}
    {
      \begin{threeparttable}
        \begin{tabularx}{\textwidth}{M{1.5cm}M{1cm}M{0.7cm}P{11cm}}
          \toprule
            \textbf{Dataset} & \textbf{Domain} & \textbf{classes} & \textbf{Example}\\
            \midrule
            Yelp-5 & restaurant & 5 & ``dr. goldberg offers everything i look for in a general practitioner. he's nice and easy to talk to without being patronizing; he's always on time in seeing his patients; he's affiliated with a top-notch hospital (nyu) which my parents have explained to me is very important in case something happens and you need surgery; and you can get referrals to see specialists without having to see him first. really, what more do you need? i'm sitting here trying to think of any complaints i have about him, but i'm really drawing a blank.''    \textit{positive++}\\
            \midrule
            SST-5 & movie & 5 & ``unlike the speedy wham-bam effect of most hollywood offerings , character development -- and more importantly , character empathy -- is at the heart of italian for beginners''    \textit{positive++}\\ 
            \midrule
            SST-2 & movie & 2 & ``jason x is positively anti-darwinian : nine sequels and 400 years later , the teens are none the wiser and jason still kills on auto-pilot ''  \textit{negative}\\ 
            \midrule
            Amazon-5 & product & 5 & ``nice screen for a nice price but..... i compared a few different flat panels with review before i narrowed down my pick, which ended up with the sylvania as over well liked. the picture got great reviews which yes it does have a good picture to look at but there are other important qualities you enjoy that makes viewing tv all the better. for example: sound... how was that forgotten?in this flat panel, it was. what a disappointment. if this is consider stereo than why does it sound like its coming from a tin can with no base at all. then too boot, if you play the dvd, the sound drops and you have to really turn up the volume to hear.i want the whole package deal: space saving, great picture, and good sound. i want to enjoy the whole experience of watching and listening. how about you?''  \textit{positive}\\
            \midrule
            Amazon-2 & product & 2 & ``not an ultimate guide. firstly,i enjoyed the format and tone of the book (how the author addressed the reader). however, i did not feel that she imparted any insider secrets that the book promised to reveal. if you are just starting to research law school, and do not know all the requirements of admission, then this book may be a tremendous help. if you have done your homework and are looking for an edge when it comes to admissions, i recommend some more topic-specific books. for example, books on how to write your personal statment, books geared specifically towards lsat preparation (powerscore books were the most helpful for me), and there are some websites with great advice geared towards aiding the individuals whom you are asking to write letters of recommendation. yet, for those new to the entire affair, this book can definitely clarify the requirements for you.'' \textit{negative}\\
            \midrule
            Sentihood & neighborhood & 2  & ``a friend of mine lived in location1 and she liked it, though other people have told me it's a bit rough'' \textit{negative}\\
            \midrule
            SemEval$_r$ & restaurant & 3 & ``if you've ever been along the river in weehawken you have an idea of the top of view the chart house has to offer'' \textit{positive}\\
            \midrule
            SemEval$_l$ & labtop & 3 & ``so if anyones looking to buy a computer or laptop you should stay far far away from any that have the name toshiba on it'' \textit{negative}\\
          \bottomrule
        \end{tabularx}
      \end{threeparttable}
    }
    \caption{\label{table:dataset_examples} Detailed information about sentiment datasets, including domain, number of classes and a concrete example}
\end{table*}

\end{document}